\documentclass[conference]{IEEEtran}
\usepackage{iftex}
  \usepackage{times}
\usepackage[numbers]{natbib}
\usepackage{multicol}
\usepackage[bookmarks=true]{hyperref}
\hypersetup{pdfauthor={Yong Hwi Song, Keon Myung Lee}, pdftitle={Planners Exploit Deterministic Error: Dithered Low-Bit World Models}, pdfsubject={Robot learning}, pdfkeywords={World models; Model-predictive control; Quantization; Stochastic rounding}, pdfcreator={}}
\usepackage{amsmath,amssymb}
\usepackage{graphicx}
\usepackage{booktabs}
\usepackage{array}
\usepackage{xspace}

\newcommand{\dtm}{\textsf{det}\xspace}
\newcommand{\shiter}{\textsf{sh-iter}\xspace}
\newcommand{\shstep}{\textsf{sh-step}\xspace}
\newcommand{\shep}{\textsf{sh-ep}\xspace}
\newcommand{\detshift}{\textsf{det+shift}\xspace}
\newcommand{\ind}{\textsf{ind}\xspace}
\newcommand{\extplan}{Sec.~\ref{sec:limits} lists the extensions we plan}
\newcommand{\extrun}{run after the main study}
\newcommand{\extsection}{Limitations and Planned Extensions}
\newcommand{\extcharter}{Planned extensions}
\newcommand{\extlistfor}{as a planned extension}

\begin{document}

\title{Planners Exploit Deterministic Error:\\Dithered Low-Bit World Models}

\author{\IEEEauthorblockN{Yong Hwi Song}
\IEEEauthorblockA{School of Information and Communication Engineering\\
Chungbuk National University\\
Email: thddydgnl1937@gmail.com}
\and
\IEEEauthorblockN{Keon Myung Lee}
\IEEEauthorblockA{School of Software\\
Chungbuk National University\\
Email: kmlee@cbnu.ac.kr}}

\maketitle

\begin{abstract}
Sampling-based planners such as MPPI and CEM query a learned world model thousands of times per control step, so low-bit world models are an obvious way to make model-based control cheaper on a robot. With round-to-nearest activation quantization and no calibration, we find that planning can collapse, by up to 100 points of success, while the world model's latent one-step prediction error stays no larger than that of a stochastic rounding scheme that plans well. At the same bit width, what decides the outcome is not the size of the rounding error but whether it persists across planner iterations and is shared across candidates. Round-to-nearest leaves an error that persists, and an iterating planner drifts toward the candidates that this error favors; in TD-MPC2, holding the same dither fixed within a step or an episode brings the drift back. Dither shared by all candidates and redrawn at every iteration removes most of the damage at the same bit width and without training: at least half of it in all 19 affected Meta-World cells (task $\times$ bit width) of TD-MPC2 (median 95\%), in 4 of 5 DINO-WM cells, and in all 8 affected tasks of a 317M-parameter multi-task TD-MPC2. Captured as CUDA graphs on an Ampere GPU, a W8A4 integer TD-MPC2 world model runs its planning rollouts 1.9--3.3$\times$ and the whole planning step 1.3--1.5$\times$ faster than FP32, though not faster than BF16 with pre-converted weights; shared dither adds at most 2.3\% latency. In TD-MPC2, block-scaled NVFP4 removes most of the damage and MXFP4 part of it, which ties the failure to coarse grids with a shared scale. Hypotheses, thresholds and analysis code were fixed before the data that test them, and the hypotheses that failed are reported.
\end{abstract}

\IEEEpeerreviewmaketitle

\section{Introduction}
\label{sec:intro}

A sampling-based planner evaluates its world model candidates $\times$ iterations $\times$ horizon times per decision: $512\times6\times3$ forward calls for the MPPI planner of TD-MPC2~\cite{hansen2024tdmpc2} and $300\times30\times5$ for the CEM planner of DINO-WM on PushT~\cite{zhou2025dinowm}. Network evaluations therefore dominate the cost of each control step, and low-bit arithmetic is a direct way to cut it (Sec.~\ref{sec:deploy}). Yet quantized world models can break planning in ways that bit width and offline prediction error do not explain~\cite{fu2026wmquant, ranganath2026bits, xu2026prede}, unless a calibration objective preserves what planning needs~\cite{anon2026modulemse}. With round-to-nearest activations and no calibration, we see planning collapse by up to 100 points of success, while in most affected cells (task $\times$ bit width) the latent one-step error of the world model is no larger than that of a stochastic rounding scheme that plans well.

We show that, at the same bit width, what decides whether a low-bit world model plans well is not the size of its rounding error but whether the error persists across planner iterations and is shared across candidates. Round-to-nearest (\dtm) is a fixed function of its input, so the error it adds to the predicted return of a given action sequence is the same in every planner iteration. Iterating planners are known to drive their candidates to where the model can no longer resolve them~\cite{anon2026elitecert, anon2026resolution}; with a persistent error, they keep the candidates that the error favors and move their sampling distribution toward them. Dither, i.e.\ stochastic rounding $\Delta\lfloor x/\Delta+u\rfloor$ with $u\sim U[0,1)$, is unbiased; its error is somewhat larger than that of round-to-nearest, but it can be redrawn. Our fix, \shiter, shares one dither draw across all candidates and redraws it at every forward call, hence at every iteration. Sharing keeps the comparison between candidates fair within an iteration, as with common random numbers~\cite{kleinman1999crn}, and redrawing denies the planner an error it can exploit across iterations.

The ingredients are classical: the optimizer's curse~\cite{smith2006curse}, the bias of sample-average approximation~\cite{kleywegt2002saa}, common random numbers, and unbiased stochastic rounding for iterative computation~\cite{gupta2015limited, croci2022sr}. Stochastic rounding is also used at inference in language-model serving, to keep error from accumulating in recurrent state caches~\cite{nvidia2026nemotron, khilar2026dither}. To our knowledge it has not been used to protect sampling-based planning, where its role is different: to deny the planner a persistent error to exploit. What is new is this use inside closed-loop planners, a control that separates sharing from persistence at a fixed per-entry error distribution, and a training-free fix whose effect we measure across two model families, two planners and two model sizes. Our contributions are:
\begin{enumerate}
\item \textbf{Phenomenon} (Sec.~\ref{sec:fix}). At the same bit width, \dtm causes planning damage that the latent one-step error does not reveal: in 37 of 44 affected cells its error is no larger than that of \shiter.
\item \textbf{Mechanism} (Sec.~\ref{sec:mech}). The planner exploits error that persists across its iterations. In TD-MPC2, positive controls that hold the same dither fixed within a step or an episode reproduce 70--130\% of \dtm's drift, and the benefit of \shiter grows with the number of iterations.
\item \textbf{Prescription} (Sec.~\ref{sec:fix}). \shiter removes at least half of the damage in all 19 affected Meta-World cells of TD-MPC2, in 4 of 5 DINO-WM cells and in all 8 affected tasks of a 317M multi-task TD-MPC2, and it is never harmful there.
\item \textbf{Deployment} (Sec.~\ref{sec:deploy}). Shared dither adds at most 2.3\% to the planning rollouts of an integer TD-MPC2 world model and under 1\% to the whole planning step, which the integer path makes 1.3--1.5$\times$ faster than FP32 but not faster than BF16 with pre-converted weights; in TD-MPC2, block-scaled FP4 formats mark the boundary of the failure.
\end{enumerate}
Every test was fixed in a hashed document before its data, and the tests that failed are reported (Table~\ref{tab:hyp}); \extplan.

\section{Related Work}
\label{sec:related}

\textbf{Quantized world models and policies.} Post-training quantization of DINO-WM makes the planning objective and task success diverge at aggressive bit widths~\cite{fu2026wmquant}, and success depends on which modules keep their precision~\cite{ranganath2026bits}; offline action deviation predicts the closed-loop loss of quantized world action models~\cite{xu2026prede}. A concurrent submission recovers full-precision planning with 4-bit weights through a calibration objective~\cite{anon2026modulemse}. Quantized vision-language-action and world-action models are benchmarked in closed loop~\cite{vlaquantbench2026, qvla2026, park2026qcrl} and deployed with fused low-bit kernels and CUDA graphs at about 2$\times$ speed-ups~\cite{wang2026steerquant, anon2026wamptq, anon2026loopquant}, and quantization error is known to recur across their denoising loops~\cite{anon2026loopquant}. These works calibrate, allocate bits or train; none changes the rounding rule or separates how the error is shared from how long it persists. Calibration and dither act on different properties of the error (Sec.~\ref{sec:limits}).

\textbf{Prediction error versus decisions.} Data-averaged prediction error does not track planning quality, while decision-relevant error does~\cite{you2026predictability, wang2026drpe}, and the signed component of rollout error along the goal direction matters more than its magnitude~\cite{anon2026magnitude}. \citet{min2026lowerror} split end-point error across a candidate pool into a pool-common and an action-relative part and repair each with an oracle. Iterating planners drive candidates to where the model cannot resolve them: certified elite decisions become rarer as CEM iterates~\cite{anon2026elitecert}, predicted candidate quality falls to chance at the frame that CEM optimizes~\cite{anon2026resolution}, and replacing early CEM rounds with a deterministic approximation hurts while score noise of the same size does not~\cite{anon2026evaluations}. We isolate which property of rounding error the planner exploits, with variants that share one per-entry error distribution, and remove it without a second model.

\textbf{Exploitation and rounding rules.} Planners exploit model error~\cite{boney2019dae}; World Models injected temperature noise into imagined rollouts so that the controller could not exploit them~\cite{ha2018worldmodels}; in CEM the world model also acts as a proposal mechanism~\cite{obst2026cem}. Dither decorrelates quantization error from the signal~\cite{roberts1962picture, schuchman1964dither}, and unbiased stochastic rounding keeps bias from accumulating~\cite{gupta2015limited, croci2022sr}. In language models, the best rounding rule differs between layers~\cite{chatelain2026sr}, a deterministic low-discrepancy dither can beat stochastic rounding in recurrent caches~\cite{khilar2026dither}, and a single fixed stochastic draw of 3-bit weights can be worse than round-to-nearest~\cite{wu2026decisions}, in line with our positive controls. These results concern error that accumulates within one computation; we study the selection bias that a planner adds when it chooses among candidates.

\section{Setup}
\label{sec:setup}

\textbf{Models and planners.} TD-MPC2~\cite{hansen2024tdmpc2}: the official 5M single-task checkpoints (three seeds) on 8 Meta-World tasks~\cite{yu2019metaworld} (success rate) and 4 DMC tasks~\cite{tassa2018dmc} (return), planned with the default MPPI~\cite{williams2017mppi} (512 samples including 24 policy trajectories, 64 elites, 6 iterations, horizon 3); and the official 317M multi-task checkpoint (mt80; distilled and quantized to FP16 in~\cite{kuzmenko2025tdmpcopt}) on the same 8 Meta-World tasks. DINO-WM~\cite{zhou2025dinowm}: the official PushT, Wall and PointMaze models with open-loop CEM~\cite{rubinstein1999cem} (300 samples, top 30, horizon 5, 30 iterations on PushT and 10 otherwise). All planner settings are the official defaults unless an experiment varies them.

\textbf{Quantization.} We quantize hidden activations: the four hidden layers of the dynamics and reward MLPs in TD-MPC2 (after Mish) and the six MLP outputs of the DINO-WM predictor (after GELU), each to $k$ bits on a uniform grid $[\min, q_{99.99}]$ calibrated per layer on held-out data. Inputs, weights and all other modules stay in floating point except in the integer kernels of Sec.~\ref{sec:deploy}. Per task, a rule fixed in advance picks the primary $k$ on held-out episodes as the largest $k$ at which \dtm loses at least 20 points of success (20\% of return on DMC), and we test $k-1$, $k$ and $k+1$: 36 cells for TD-MPC2 at 5M and 9 for DINO-WM. For each 317M task we test the primary $k$ (3--5 bits).

\textbf{Rounding variants.} All stochastic variants use the same unbiased dither and therefore the same per-entry error distribution; they differ only in how $u$ is shared and how long it is kept. \ind draws $u$ independently per candidate at every forward call. \shiter shares one draw per layer across all candidates and redraws it at every forward call. \shstep and \shep share the same draw but hold it fixed within a control step or an episode (they coincide in open-loop CEM, where we call them sh-plan). \detshift keeps round-to-nearest but shifts the shared grid by a fresh offset at every call.

\textbf{Metrics and statistics.} Every variant is paired with the unquantized model on the same episode seeds and planner noise (90 pairs per cell for TD-MPC2: three checkpoints $\times$ 30 seeds, or one 317M checkpoint $\times$ 90 seeds; 100 for DINO-WM). The damage $D$ is the drop from that baseline in success points (relative return on DMC). A cell is \emph{affected} if the \dtm drop is real under a one-sided paired bootstrap ($10^4$ resamples) with Holm correction over the cells of a task. The fraction of damage removed by a variant $v$ is $\mathrm{red}=(D_{\dtm}-D_v)/D_{\dtm}$, and $v$ is \emph{harmful} if the 95\% interval of $D_v-D_{\dtm}$ lies above 5 points. The latent one-step error is the relative error of the predicted next latent state for the first-iteration candidates at planning time.

\textbf{Exploitation.} Let $e(a)$ be the error of the quantized model's predicted return for candidate $a$ relative to the unquantized model. The optimizer exploitation index $\mathrm{OEI}=e(a_{\mathrm{exec}})-\bar e_{\mathrm{first}}$ compares the executed plan with random first-iteration candidates and splits exactly into a \emph{selection} term (the last iteration's pick against its population) and a \emph{drift} term, the change in the population-mean error from the first to the last iteration. Negative drift means the planner moved its population toward candidates that the error favors.

\textbf{Analysis plan and compute.} The study ran in stages, from environment checks to the extensions marked in Table~\ref{tab:hyp}. Before each stage, its hypotheses, thresholds, seeds, sample sizes and analysis code were frozen in a design document whose hash was logged before any of the stage's data existed; operational choices made in smoke tests were recorded before the main runs. Analyses chosen after seeing data are labeled post hoc. The main study used about 147 GPU-hours on RTX A5000 GPUs, and the extensions used cloud A40, RTX 4090 and RTX 5090 GPUs. Code, design documents and their hashes will be released with the final paper.

\begin{table*}[t]
\centering
\footnotesize
\setlength{\tabcolsep}{4pt}
\caption{Hypotheses and outcomes. Each test was fixed before its data; rows marked $^\dagger$ were added after the first stage and tested on new seeds; $^\ddagger$ marks the extension experiments \extrun. Mechanism tasks: mw-push, mw-pick-place, cheetah-run, PushT.}
\label{tab:hyp}
\begin{tabular}{@{}l>{\raggedright\arraybackslash}p{0.43\textwidth}>{\raggedright\arraybackslash}p{0.40\textwidth}@{}}
\toprule
& Hypothesis and test & Outcome \\
\midrule
H1 & \dtm's latent one-step error $\le$ \shiter's in some affected cell of each family & holds: 29/36 cells (TD-MPC2 5M, DINO-WM); 317M$^\ddagger$ 8/8 \\
H3a & \shiter removes $\ge$50\% of the damage in $\ge$75\% of affected cells, none harmful & Meta-World 19/19 (median 95\%), DINO-WM 4/5, 317M$^\ddagger$ 8/8; DMC fails (5/12) \\
DMC$^\dagger$ & Spearman(persistent share of \dtm's error, red of \shiter) $\ge 0.5$ over 12 tasks & holds: 0.83 \\
H2a$'^\dagger$ & \dtm drift $<0$ and drift(\shiter)/drift(\dtm) $<0.5$ & holds: 4/4 tasks (ratio $\le 0.24$) \\
H2c & Spearman(candidate count, $|\mathrm{OEI}_{\dtm}|$ and $D_{\dtm}$) $\ge 0.8$, elites fixed & holds: 3/4 tasks \\
H4a & Spearman(table size of fixed dither, its damage) $\le -0.8$ & holds: 5/6 tasks (three exactly at $-0.8$) \\
H3b & \shiter beats \dtm on top of the best post-training quantization & TD-MPC2 holds (10/19); DINO-WM fails (1/3) \\
H5 & Dither latency overhead $\le$10\% in a fused integer kernel & holds (max 6.9\%); integer kernel 18\% slower than FP32 (eager) \\
L1--L3$^\ddagger$ & Integer rollout $\le$0.8$\times$ FP32 time and $\le$0.9$\times$ the faster of TF32/BF16 (autocast) in some condition; dither $\le$5\% everywhere & all hold: 12/12, 10/12, max overhead 2.1\%; replicated in a second run (2.3\%) \\
L2c$^\ddagger$ & As L2, with pre-converted BF16 among the float formats & fails: 0/12 (1.07--1.52$\times$ its time) \\
S1--S3$^\ddagger$ & As L1--L3 (with pre-converted BF16) for the whole planning step & S1, S3 hold (12/12; max 0.7\%); S2 fails (0/12, 1.01--1.12$\times$) \\
F1--F3$^\ddagger$ & FP4 formats: \dtm damages $\ge$3 of 12 tasks; \shiter removes $\ge$50\%; \dtm's error $\le$ \shiter's & NVFP4 fails F1 (2/12), MXFP4 holds (4/12); F2 holds on Meta-World (1 cell), fails on DMC (2/5); F3 holds (6/6) \\
H2a, H2b & Original OEI tests: OEI$\approx$0 under \shiter; OEI predicts damage better than one-step error & fail (claims narrowed to drift) \\
H4b & Benefit of sharing grows with the density of near-ties & fails ($\rho=0.03$) \\
\bottomrule
\end{tabular}
\end{table*}

\section{Phenomenon and Prescription}
\label{sec:fix}

\begin{figure}[t]
\centering
\includegraphics[width=\columnwidth]{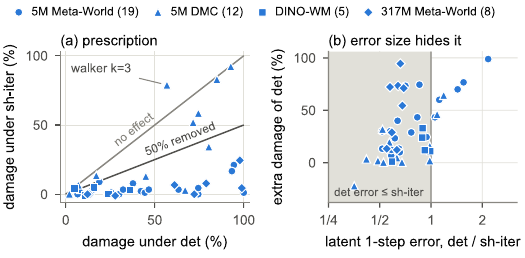}
\caption{Every affected cell. (a) Damage under \shiter vs.\ \dtm (success points on Meta-World and DINO-WM, relative return on DMC); below the dashed line \shiter removes at least half of the damage. (b) \dtm's extra damage against the ratio of the two latent one-step errors: most cells sit in the shaded region, where \dtm's error is the smaller one.}
\label{fig:damage}
\end{figure}

\textbf{Damage that the latent error does not show.} \dtm costs up to 100 points (mw-button-press at 2 bits). In 29 of the 36 affected cells of DINO-WM and the 5M TD-MPC2, \dtm's latent one-step error is no larger than that of \shiter (Fig.~\ref{fig:damage}b). On mw-pick-place at 3 bits, for example, the error is 0.22 for \dtm and 0.39 for \shiter, yet \dtm loses 37.8 points and \shiter 7.8. The 317M model shows the same pattern in all 8 affected tasks: \dtm's latent error is 0.55--0.69 times that of \shiter while \dtm loses 12--99 points. The comparison is within a bit width; across bit widths one-step error and damage do rank together. Two other error measures do separate \dtm: errors over three or five steps (PushT, 4 bits, five steps: 0.25 for \dtm vs.\ 0.04 for \shiter), and, observed after the fact, the one-step error of the predicted reward, which is larger under \dtm in all 19 affected Meta-World cells and all 8 tasks at 317M. A deterministic error is correlated across units and adds up in a scalar output, while independent dither errors partly cancel; we have not tested this explanation and list it \extlistfor\ (Sec.~\ref{sec:limits}).

\textbf{Prescription.} \shiter removes at least half of the damage in all 19 affected Meta-World cells (median 95\%) and in 4 of 5 DINO-WM cells, and it is not harmful in any of them (Fig.~\ref{fig:damage}a, Table~\ref{tab:hyp}). The stochastic variants share one per-entry error distribution, yet across the 19 affected Meta-World cells the mean drop is 50.1 points under \shep, 34.9 under \shstep, 29.2 under \ind and 4.0 under \shiter, against 44.6 under \dtm; \detshift also loses only 4.0. The weakest Meta-World cell keeps 47\% of the damage (window-open, 4 bits), and the DINO-WM exception is PushT at 5 bits, where \dtm's damage is only 5 points. On the 317M multi-task model, which needs no change to the method, \shiter removes 75--104\% of the damage in all 8 tasks (e.g.\ mw-peg-insert-side: 98.9 points lost under \dtm, 4.4 under \shiter), and the 95\% interval of $D_{\shiter}-D_{\dtm}$ lies below zero in every task. The primary bit widths at 317M (3--5) are close to those at 5M (3--4), so the larger model is not more robust to rounding.

\textbf{Where it helps less.} On DMC, \shiter removes half of the damage in only 5 of 12 cells and is harmful on walker at 3 bits (78.9\% of return lost vs.\ 56.9\% under \dtm). A rule added after the first stage and tested on new seeds explains where it helps: the persistent share of \dtm's favorable error (drift over OEI) is 0.33--0.52 on DMC and 0.54--0.73 on Meta-World, and across the 12 TD-MPC2 tasks it predicts the damage removed (Spearman 0.83, new seeds). The two DMC tasks with the smallest persistent share are the two where \shiter fails.

\section{Mechanism}
\label{sec:mech}

\begin{figure*}[t]
\centering
\includegraphics[width=\textwidth]{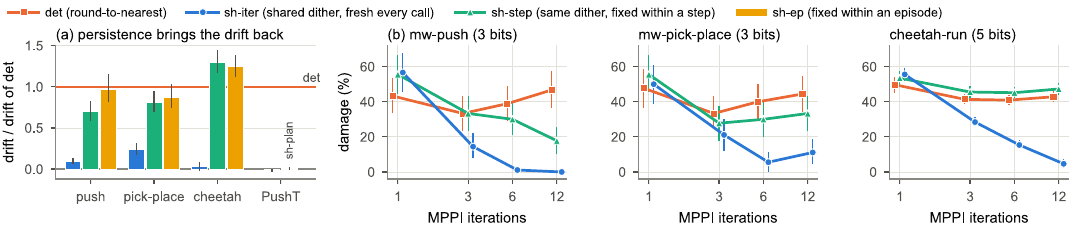}
\caption{Mechanism. (a) Favorable drift of the population-mean error kept by each variant, relative to \dtm (TD-MPC2 at its primary bit width; PushT at 4 bits, where \shstep and \shep coincide as sh-plan); bars show 95\% bootstrap intervals. (b) Damage vs.\ MPPI iterations (TD-MPC2, 90 paired episodes per point, 95\% intervals). Fresh dither needs iterations to help; \dtm does not improve with them.}
\label{fig:mech}
\end{figure*}

\textbf{Drift.} Under \dtm the population-mean error drifts in the favorable direction in all four mechanism tasks at their primary bit widths. \shiter keeps at most 24\% of that drift (0.10, 0.24, 0.03 and $-0.01$ of it), and \detshift keeps at most 15\%. The positive controls isolate persistence: holding the \emph{same} shared dither fixed for a control step (\shstep) or an episode (\shep) brings back 70--130\% of \dtm's drift in the three TD-MPC2 tasks (Fig.~\ref{fig:mech}a). Error size and sharing are equal across these variants; only persistence differs. On PushT the full-size fixed table did not reproduce the drift (ratio 0.01); there a fixed dither was exploited only when we made its table of offsets coarse, as below.

\textbf{Iterations, candidates and table size.} The fix needs an iterating planner. With one MPPI iteration \shiter is no better than \dtm; at 12 iterations its damage is 0--11\% while \dtm's stays at 43--47\% and \shstep's at 18--47\% (Fig.~\ref{fig:mech}b). With the number of elites fixed, \dtm's exploitation $|\mathrm{OEI}|$ and damage grow with the number of candidates in three of four tasks, the dose-response expected from the optimizer's curse~\cite{smith2006curse}. A fixed dither drawn from a table of $N_{\mathrm{eff}}$ independent offsets hurts less as the table grows (five of six tasks, three of them exactly at the threshold), and in an exploratory Wall cell (3 bits) a single offset per layer collapsed planning (+37 points).

\textbf{Theory.} For a planner that averages fresh unbiased noise, the expected drift is zero. A persistent error field instead pulls the population toward the minimizer of the smoothed model cost, the bias of sample-average approximation~\cite{kleywegt2002saa}. The selection and drift terms we derived matched 922k simulated MPPI and CEM runs (fitted slopes 0.92--1.05) and predicted, before the first stage was analyzed, that the OEI would be negative even under fresh noise, because the last iteration's pick benefits from the noise it was selected on (the winner's curse). This is why our original OEI tests, H2a and H2b, failed, and why our mechanism claim concerns the drift rather than the OEI.

\section{Deployment: Latency and Number Formats}
\label{sec:deploy}

\begin{figure}[t]
\centering
\includegraphics[width=\columnwidth]{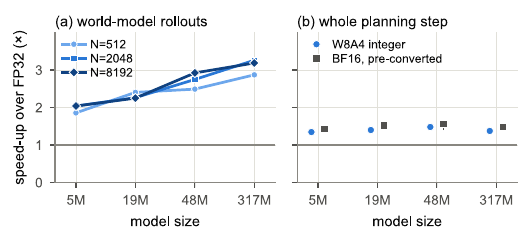}
\caption{Speed-up over FP32 on an NVIDIA A40, as CUDA graphs. (a) W8A4 world-model rollouts of one MPPI step ($N$ candidates, 6 iterations, horizon 3). (b) Whole planning step, W8A4 and BF16 with pre-converted weights: median over $N$, bars show the range. (a) and (b) are two runs on different A40s.}
\label{fig:latency}
\end{figure}

\textbf{Integer kernels.} A fused Triton kernel (int8 weights, 4- or 8-bit activation codes, shared dither drawn on the GPU) reproduces the fake-quantization results in closed loop, and 8-bit activations cause no damage. Run eagerly at 5M parameters and 512 candidates, it was 18\% slower than FP32: the step is bound by kernel launches, not arithmetic. To map where low bit widths pay off, we timed the world-model rollouts of one control step (dynamics and reward for $N$ candidates, plus the policy trajectories) replayed as a single CUDA graph, as TD-MPC2's compiled planner does. The integer path uses int8 weights (per output channel) and 4-bit activation codes with int8 tensor-core GEMMs; the first layer, whose input $[z,a]$ is not quantized, stays in FP32. On an NVIDIA A40, across 5M--317M parameters and $N\in\{512, 2048, 8192\}$ (10 interleaved rounds per condition), the integer rollout is 1.9--3.3$\times$ faster than FP32 in all 12 conditions (Fig.~\ref{fig:latency}a); shared dither changes latency by $-0.2$ to $+2.1\%$, because one draw per layer and call serves every candidate. The gains are largest at 317M, the size at which the multi-task results of Sec.~\ref{sec:fix} hold. A second run on another A40 reproduced these speed-ups (1.9--3.3$\times$; dither at most $+2.3\%$) and tested a stronger BF16 baseline, with the weights converted once in advance instead of at every call (autocast): it is then the fastest float format in all 12 conditions, and our integer rollout takes 1.07--1.52$\times$ its time. Much of the gap is the first layer, which we keep in FP32; with TF32 there, the integer rollout takes 0.72--0.96$\times$ the time of pre-converted BF16 (post hoc), but we have not evaluated that path in closed loop. Because the dynamics and reward networks are only 40--52\% of an FP32 planning step, the whole step (encoder, policy, Q-ensemble and MPPI captured as one CUDA graph) gains less: 1.28--1.49$\times$ over FP32 (Fig.~\ref{fig:latency}b), while taking 1.01--1.12$\times$ the time of the step with pre-converted BF16; shared dither adds at most 0.7\% to it.

\textbf{FP4 formats mark the boundary.} On an RTX 5090, an NVFP4 GEMM at the 317M layer width is up to 5.3$\times$ faster than BF16, so we repeated the 5M experiment with activations in NVFP4~\cite{nvidia2025nvfp4} (blocks of 16, E4M3 scale) and MXFP4~\cite{rouhani2023mx} (blocks of 32, power-of-two scale), rounding each block to the E2M1 grid to nearest (\dtm) or stochastically (\shiter). \dtm damaged only 2 of 12 tasks under NVFP4 (cheetah-run 9.9\%, walker 1.7\%), short of the three our analysis plan required, and 4 of 12 under MXFP4 (cheetah-run 31.3\%, window-open 10.0 points), against 71.6\% on cheetah-run with a uniform 4-bit grid. Where damage remained, the signature was the same: \dtm's latent one-step error was 0.48--0.63 times that of \shiter in all six damaged cells, and \shiter removed most of the damage in the one damaged Meta-World cell but in only 2 of 5 DMC cells, as with uniform grids. Block scaling shrinks the error and dither removes its persistence, two different axes; in TD-MPC2 the failure we study belongs to coarse grids with a shared scale, while quantized VLA policies can still lose success under NVFP4~\cite{park2026qcrl}.

\section{\extsection}
\label{sec:limits}

\textbf{Limitations.} The prescription assumes an iterating planner, and on DMC it helps in fewer than half of the affected cells. On top of the best post-training quantization (mostly per-channel ranges), which already removes most of \dtm's damage, the extra gain of \shiter is borderline in TD-MPC2 (10 of 19 cells) and absent in DINO-WM (1 of 3). We did not combine dither with calibration objectives such as~\cite{anon2026modulemse}. Our quantization-aware-training baseline was weak, so we draw no conclusion about training. The 317M results come from one checkpoint and the FP4 results from fake quantization of the 5M models. The latency map comes from one GPU type, on which BF16 with pre-converted weights, not evaluated in closed loop, is faster than our integer path with its first layer in FP32. We have no physical-robot experiments.

\textbf{\extcharter.} As in Table~\ref{tab:hyp}, each item would have its hypotheses, thresholds and analysis code fixed before its data.
\begin{enumerate}
\item \emph{A diagnostic that sees the damage.} Unlike the latent one-step error, the one-step reward error separated \dtm in every affected Meta-World cell (post hoc). We will test on new tasks and seeds whether the correlation of rounding error across hidden units predicts \dtm's damage within a bit width, and whether a reward-error check before deployment flags the damaging configurations.
\item \emph{Integer kernels for DINO-WM.} We will port the integer path to the DINO-WM predictor and test whether closed-loop damage and repair match fake quantization in the affected cells, with the same latency budget for dither ($\le$5\%).
\item \emph{Closing the gap to BF16.} We will move the integer path's first layer out of FP32 and evaluate that path and BF16 with pre-converted weights in closed loop, comparing success and latency on the same planner.
\item \emph{Embedded and FP4 hardware.} We will repeat the latency map, against BF16 with pre-converted weights, on an NVIDIA Jetson module and, with fused quantization kernels, for NVFP4 on a Blackwell GPU, including the cost of drawing dither.
\end{enumerate}
The negative results above (DMC, the combination with strong post-training quantization, planners with few iterations) stay in scope as boundaries to report, not as targets to tune away.

\bibliographystyle{plainnat}
\bibliography{refs}

\begin{thebibliography}{40}
\providecommand{\natexlab}[1]{#1}
\providecommand{\url}[1]{\texttt{#1}}
\expandafter\ifx\csname urlstyle\endcsname\relax
  \providecommand{\doi}[1]{doi: #1}\else
  \providecommand{\doi}{doi: \begingroup \urlstyle{rm}\Url}\fi

\bibitem[Anonymous(2026{\natexlab{a}})]{anon2026elitecert}
Anonymous.
\newblock {When Can a World Model Decide? Certifying Elite Decisions in
  Sampling-Based Planning}.
\newblock Submitted to ICLR 2027, 2026{\natexlab{a}}.
\newblock OpenReview {cVCKr4bBU2}.

\bibitem[Anonymous(2026{\natexlab{b}})]{anon2026evaluations}
Anonymous.
\newblock {Not All Evaluations Are Equal: Where Test-Time Compute Matters in
  World-Model Planning}.
\newblock Submitted to ICLR 2027, 2026{\natexlab{b}}.
\newblock OpenReview {1vmM17t1R8}.

\bibitem[Anonymous(2026{\natexlab{c}})]{anon2026loopquant}
Anonymous.
\newblock {LoopQuant: Loop-Aware Post-Training Quantization for
  Vision-Language-Action Models}.
\newblock Submitted to ICLR 2027, 2026{\natexlab{c}}.
\newblock OpenReview {AhG6FcJBHt}.

\bibitem[Anonymous(2026{\natexlab{d}})]{anon2026magnitude}
Anonymous.
\newblock {What Rollout Error Magnitude Misses in Latent World-Model Planning}.
\newblock Submitted to ICLR 2027, 2026{\natexlab{d}}.
\newblock OpenReview {5alhlhb7zd}.

\bibitem[Anonymous(2026{\natexlab{e}})]{anon2026modulemse}
Anonymous.
\newblock {What Must Quantization Preserve in Latent World Models?}
\newblock Submitted to ICLR 2027, 2026{\natexlab{e}}.
\newblock OpenReview {3VSa1Kx7HD}.

\bibitem[Anonymous(2026{\natexlab{f}})]{anon2026resolution}
Anonymous.
\newblock {Resolution Collapse in Latent-Space CEM Planning}.
\newblock Submitted to ICLR 2027, 2026{\natexlab{f}}.
\newblock OpenReview {eTu4mzKor3}.

\bibitem[Anonymous(2026{\natexlab{g}})]{anon2026wamptq}
Anonymous.
\newblock {WAMPTQ: Beyond Weight Quantization via Task-Utility-Guided Precision
  Allocation for Efficient World Model Deployment}.
\newblock Submitted to ICLR 2027, 2026{\natexlab{g}}.
\newblock OpenReview {hdBfu8lIC7}.

\bibitem[Boney et~al.(2019)Boney, Di~Palo, Berglund, Ilin, Kannala, Rasmus, and
  Valpola]{boney2019dae}
Rinu Boney, Norman Di~Palo, Mathias Berglund, Alexander Ilin, Juho Kannala,
  Antti Rasmus, and Harri Valpola.
\newblock Regularizing trajectory optimization with denoising autoencoders.
\newblock In \emph{Advances in Neural Information Processing Systems
  (NeurIPS)}, 2019.

\bibitem[Chatelain and de~Oliveira~Castro(2026)]{chatelain2026sr}
Yohan Chatelain and Pablo de~Oliveira~Castro.
\newblock {Stochastic Rounding in Low-Precision Transformer Inference: A
  Variable-Precision Emulation Study of a Small GPT-2}.
\newblock \emph{arXiv preprint arXiv:2610.01889}, 2026.

\bibitem[Croci et~al.(2022)Croci, Fasi, Higham, Mary, and
  Mikaitis]{croci2022sr}
Matteo Croci, Massimiliano Fasi, Nicholas~J. Higham, Theo Mary, and Mantas
  Mikaitis.
\newblock Stochastic rounding: Implementation, error analysis and applications.
\newblock \emph{Royal Society Open Science}, 9\penalty0 (3):\penalty0 211631,
  2022.

\bibitem[Fu et~al.(2026)Fu, Zhao, Han, Zhou, Chen, and Wang]{fu2026wmquant}
Zhongqian Fu, Tianyi Zhao, Kai Han, Hang Zhou, Xinghao Chen, and Yunhe Wang.
\newblock {An Empirical Study of World Model Quantization}.
\newblock \emph{arXiv preprint arXiv:2602.02110}, 2026.

\bibitem[Gupta et~al.(2015)Gupta, Agrawal, Gopalakrishnan, and
  Narayanan]{gupta2015limited}
Suyog Gupta, Ankur Agrawal, Kailash Gopalakrishnan, and Pritish Narayanan.
\newblock Deep learning with limited numerical precision.
\newblock In \emph{International Conference on Machine Learning (ICML)}, pages
  1737--1746, 2015.

\bibitem[Ha and Schmidhuber(2018)]{ha2018worldmodels}
David Ha and J{\"u}rgen Schmidhuber.
\newblock Recurrent world models facilitate policy evolution.
\newblock In \emph{Advances in Neural Information Processing Systems
  (NeurIPS)}, 2018.

\bibitem[Hansen et~al.(2024)Hansen, Su, and Wang]{hansen2024tdmpc2}
Nicklas Hansen, Hao Su, and Xiaolong Wang.
\newblock {TD-MPC2}: Scalable, robust world models for continuous control.
\newblock In \emph{International Conference on Learning Representations
  (ICLR)}, 2024.

\bibitem[Khilar(2026)]{khilar2026dither}
Snigdha~Chandan Khilar.
\newblock {Low-Discrepancy Dither for Quantized Recurrent State Caches}.
\newblock \emph{arXiv preprint arXiv:2609.39185}, 2026.

\bibitem[Kleinman et~al.(1999)Kleinman, Spall, and Naiman]{kleinman1999crn}
Nathan~L. Kleinman, James~C. Spall, and Daniel~Q. Naiman.
\newblock Simulation-based optimization with stochastic approximation using
  common random numbers.
\newblock \emph{Management Science}, 45\penalty0 (11):\penalty0 1570--1578,
  1999.

\bibitem[Kleywegt et~al.(2002)Kleywegt, Shapiro, and Homem-de
  Mello]{kleywegt2002saa}
Anton~J. Kleywegt, Alexander Shapiro, and Tito Homem-de Mello.
\newblock The sample average approximation method for stochastic discrete
  optimization.
\newblock \emph{SIAM Journal on Optimization}, 12\penalty0 (2):\penalty0
  479--502, 2002.

\bibitem[Kuzmenko and Shvai(2025)]{kuzmenko2025tdmpcopt}
Dmytro Kuzmenko and Nadiya Shvai.
\newblock {TD-MPC-Opt: Distilling Model-Based Multi-Task Reinforcement Learning
  Agents}.
\newblock \emph{arXiv preprint arXiv:2507.01823}, 2025.

\bibitem[Min et~al.(2026)Min, Li, Xu, and Du]{min2026lowerror}
Rui Min, Xianyao Li, Fang Xu, and Jing Du.
\newblock {When Low Prediction Error Misleads Planning: Diagnosing
  Representation, Dynamics, and Decision Failures in Latent World Models}.
\newblock \emph{arXiv preprint arXiv:2610.05550}, 2026.

\bibitem[{NVIDIA}(2026)]{nvidia2026nemotron}
{NVIDIA}.
\newblock {Nemotron 3 Super: Open, Efficient Mixture-of-Experts Hybrid
  Mamba-Transformer Model for Agentic Reasoning}.
\newblock \emph{arXiv preprint arXiv:2604.12374}, 2026.

\bibitem[{NVIDIA} et~al.(2025){NVIDIA}, Abecassis, Agrusa,
  et~al.]{nvidia2025nvfp4}
{NVIDIA}, Felix Abecassis, Anjulie Agrusa, et~al.
\newblock Pretraining large language models with {NVFP4}.
\newblock \emph{arXiv preprint arXiv:2509.25149}, 2025.

\bibitem[Obst and Stolzenburg(2026)]{obst2026cem}
Oliver Obst and Frieder Stolzenburg.
\newblock {In CEM, a World Model Is Also a Proposal Mechanism}.
\newblock \emph{arXiv preprint arXiv:2610.00921}, 2026.

\bibitem[Park et~al.(2026)Park, Lee, Heo, et~al.]{park2026qcrl}
Seongmin Park, Jaewon Lee, Yujun Heo, et~al.
\newblock {Recovering Low-Bit Failures in Vision-Language-Action Policies via
  Quantization-Conditioned Reinforcement Learning}.
\newblock \emph{arXiv preprint arXiv:2610.15675}, 2026.

\bibitem[Ranganath et~al.(2026)Ranganath, Patnaik, and
  Menon]{ranganath2026bits}
Suraj Ranganath, Anish Patnaik, and Vaishak Menon.
\newblock {Where Bits Matter in World Model Planning: A Paired Mixed-Bit Study
  for Efficient Spatial Reasoning}.
\newblock \emph{arXiv preprint arXiv:2602.11882}, 2026.

\bibitem[Roberts(1962)]{roberts1962picture}
Lawrence~G. Roberts.
\newblock Picture coding using pseudo-random noise.
\newblock \emph{IRE Transactions on Information Theory}, 8\penalty0
  (2):\penalty0 145--154, 1962.

\bibitem[Rouhani et~al.(2023)Rouhani, Zhao, More, et~al.]{rouhani2023mx}
Bita~Darvish Rouhani, Ritchie Zhao, Ankit More, et~al.
\newblock Microscaling data formats for deep learning.
\newblock \emph{arXiv preprint arXiv:2310.10537}, 2023.

\bibitem[Rubinstein(1999)]{rubinstein1999cem}
Reuven~Y. Rubinstein.
\newblock The cross-entropy method for combinatorial and continuous
  optimization.
\newblock \emph{Methodology and Computing in Applied Probability}, 1\penalty0
  (2):\penalty0 127--190, 1999.

\bibitem[Schuchman(1964)]{schuchman1964dither}
Leonard Schuchman.
\newblock Dither signals and their effect on quantization noise.
\newblock \emph{IEEE Transactions on Communication Technology}, 12\penalty0
  (4):\penalty0 162--165, 1964.

\bibitem[Smith and Winkler(2006)]{smith2006curse}
James~E. Smith and Robert~L. Winkler.
\newblock The optimizer's curse: Skepticism and postdecision surprise in
  decision analysis.
\newblock \emph{Management Science}, 52\penalty0 (3):\penalty0 311--322, 2006.

\bibitem[Tassa et~al.(2018)Tassa, Doron, Muldal, Erez, Li, de~Las~Casas,
  Budden, Abdolmaleki, Merel, Lefrancq, Lillicrap, and
  Riedmiller]{tassa2018dmc}
Yuval Tassa, Yotam Doron, Alistair Muldal, Tom Erez, Yazhe Li, Diego
  de~Las~Casas, David Budden, Abbas Abdolmaleki, Josh Merel, Andrew Lefrancq,
  Timothy Lillicrap, and Martin Riedmiller.
\newblock {DeepMind} control suite.
\newblock \emph{arXiv preprint arXiv:1801.00690}, 2018.

\bibitem[Wang et~al.(2026{\natexlab{a}})Wang, Fan, and Huang]{wang2026drpe}
Linhao Wang, Yiyan Fan, and Dongjin Huang.
\newblock {Not All Errors Matter: Decision-Relevant Prediction Error Predicts
  Planning Quality}.
\newblock \emph{arXiv preprint arXiv:2609.32322}, 2026{\natexlab{a}}.
\newblock v2.

\bibitem[Wang et~al.(2026{\natexlab{b}})Wang, Wang, Liu,
  et~al.]{wang2026steerquant}
Yunhan Wang, Haodong Wang, Zhiming Liu, et~al.
\newblock {SteerQuant: Steering Quantization Error with Action-Guided Scaling
  in World-Action Models}.
\newblock \emph{arXiv preprint arXiv:2609.39056}, 2026{\natexlab{b}}.

\bibitem[Williams et~al.(2017)Williams, Wagener, Goldfain, Drews, Rehg, Boots,
  and Theodorou]{williams2017mppi}
Grady Williams, Nolan Wagener, Brian Goldfain, Paul Drews, James~M. Rehg, Byron
  Boots, and Evangelos~A. Theodorou.
\newblock Information theoretic {MPC} for model-based reinforcement learning.
\newblock In \emph{IEEE International Conference on Robotics and Automation
  (ICRA)}, pages 1714--1721, 2017.

\bibitem[Wu et~al.(2026)Wu, Dhiman, and Koshiyama]{wu2026decisions}
Zekun Wu, Swati Dhiman, and Adriano Koshiyama.
\newblock {Which Decisions Low-Bit Quantization Breaks, and How to Predict
  Them}.
\newblock \emph{arXiv preprint arXiv:2608.06564}, 2026.

\bibitem[Xu et~al.(2026{\natexlab{a}})Xu, Guo, Chen, Du, and Shi]{xu2026prede}
Jiuyi Xu, Jinjia Guo, Meida Chen, Jing Du, and Yangming Shi.
\newblock {Predict Before You Deploy: Offline Prediction of
  Quantization-Induced Task Degradation for World Action Models}.
\newblock \emph{arXiv preprint arXiv:2609.19441}, 2026{\natexlab{a}}.

\bibitem[Xu et~al.(2026{\natexlab{b}})Xu, Jin, Chen, Wang, Sui, and
  Shi]{vlaquantbench2026}
Jiuyi Xu, Qing Jin, Meida Chen, Song Wang, Yang Sui, and Yangming Shi.
\newblock {VLAQuantBench: Closed-Loop Evaluation of Post-Training Quantization
  for Vision-Language-Action Models}.
\newblock \emph{arXiv preprint arXiv:2609.25376}, 2026{\natexlab{b}}.

\bibitem[Xu et~al.(2026{\natexlab{c}})Xu, Yang, Fan, et~al.]{qvla2026}
Yuhao Xu, Yantai Yang, Zhenyang Fan, et~al.
\newblock {QVLA: Not All Channels Are Equal in Vision-Language-Action Model's
  Quantization}.
\newblock In \emph{International Conference on Learning Representations
  (ICLR)}, 2026{\natexlab{c}}.

\bibitem[You et~al.(2026)You, Zhang, Ran, et~al.]{you2026predictability}
Hanzhe You, Yonggang Zhang, Maohao Ran, et~al.
\newblock {A Control Theory of Predictability in Latent World Models}.
\newblock \emph{arXiv preprint arXiv:2607.10362}, 2026.

\bibitem[Yu et~al.(2019)Yu, Quillen, He, Julian, Hausman, Finn, and
  Levine]{yu2019metaworld}
Tianhe Yu, Deirdre Quillen, Zhanpeng He, Ryan Julian, Karol Hausman, Chelsea
  Finn, and Sergey Levine.
\newblock Meta-world: A benchmark and evaluation for multi-task and meta
  reinforcement learning.
\newblock In \emph{Conference on Robot Learning (CoRL)}, 2019.

\bibitem[Zhou et~al.(2024)Zhou, Pan, LeCun, and Pinto]{zhou2025dinowm}
Gaoyue Zhou, Hengkai Pan, Yann LeCun, and Lerrel Pinto.
\newblock {DINO-WM}: World models on pre-trained visual features enable
  zero-shot planning.
\newblock \emph{arXiv preprint arXiv:2411.04983}, 2024.

\end{thebibliography}

\end{document}